%% file: main.tex
\documentclass[letterpaper, 10 pt, conference]{ieeeconf}  % 

\IEEEoverridecommandlockouts 

\usepackage{amsmath}
\usepackage{hyperref}
\usepackage{multirow}
\usepackage{graphicx}
\usepackage{subcaption}
\usepackage{caption}

\usepackage{xcolor}

\title{\LARGE \bf
Fast Enough to Act: Spatio-Temporal Visual Token Merging for Low-Latency Robotic VLMs and VLAs
}

\author{Junzhou Chen$^{1}$, Jindong Wang$^{2}$, Gang Zhou$^{1}$% <-this % stops a space
\thanks{
$^{1}$J. Chen and G. Zhou are with the Department of Computer Science, and
$^{2}$J. Wang is with the Department of Data Science,
William \& Mary, Williamsburg, VA 23188, USA.
Emails: \{jchen57, jdw, gzhou\}@wm.edu.
}}%
\begin{document}

\maketitle
\thispagestyle{empty}
\pagestyle{empty}

\begin{abstract}

Vision-language models and vision-language action models endow the robot with unprecedented capabilities.
However, the input of video and high-resolution images yields a massive number of visual tokens, leading to extremely high inference latency and severely hindering the robot's real-time control.
To break through this computational bottleneck, we propose ST-Merge, a plug-and-play, training-free framework that efficiently fuses redundant tokens directly during the visual encoding phase.
By explicitly constructing 3D spatiotemporal coordinates, it employs a multi-queue parallel matching and weighted aggregation mechanism to achieve efficient and geometrically consistent fusion of redundant tokens across frames.
In addition, we introduce a post-merge positional correction mechanism that effectively eliminates spatial deviation caused by merging by dynamically re-evaluating the rotational position code of the weighted centroid of the vision token, thereby ensuring the high-precision spatial awareness required for dexterous operation.
In the Video Question Answering task on the mainstream VLM, Qwen2.5-VL, ST-Merge achieves a 2$\times$ inference speedup with only a tiny 1\% loss in precision.
When deployed on the $\pi_{0.5}$ VLA policy, ST-Merge achieves an 8.3$\times$ speedup at 1024 $\times$ 1024 resolution and matches the baseline success rate at this high-resolution setting. At lower resolutions, it introduces a small drop in accuracy.
Code is available at: \url{https://github.com/Junzhou-Chen/ST_Merge}

\end{abstract}

\input{Content/01_Intro}

\input{Content/02_Related_Work}

\input{Content/03_Method}
\input{Content/04_Experiment}

\input{Content/05_Conclusion}

\bibliographystyle{IEEEtran}
\bibliography{Ref/IEEEexample}

\end{document}

%% file: Content/01_Intro.tex
\section{Introduction}
\label{sec:introduction}
Recently, with the rapid development of Large Language Models (LLMs)~\cite{grattafiori2024llama, achiam2023gpt, liu2024deepseek} and Vision-Language Models (VLMs)~\cite{bai2025qwen2, yang2025qwen3, li2024llava}, the intellectual ability of robotic systems has significantly improved, enabling them to understand and process visual scenes and complex instructions. 
Resulting in generalization performance that was previously difficult to achieve. 
These models have shown considerable benefits across domains, from path planning~\cite{batool2025impedancegpt}, visual semantic localization~\cite{zhu2025rod}, and risk-aware reasoning~\cite{wu2024safety}.

To leverage the superior capabilities of VLMs for robotic manipulation and closed-loop decision-making, recent research has extended VLMs into Vision Language Action Models (VLAs)~\cite{kim2024openvla, intelligence2025pi_}  by mapping VLM semantics to robot action policy.
VLAs transform high-level observations and decisions into commands that robots can directly manipulate.
Compared with traditional methods~\cite{chi2025diffusion}, which rely on task-specific representations, VLAs demonstrate better generalization and better cross-distribution transfer when processing unseen commands and long-range tasks.

\begin{figure}[t]
    \centering
    
    % -------- (a) ----------
    \begin{subfigure}{\columnwidth}
        \centering
        \includegraphics[width=\columnwidth]{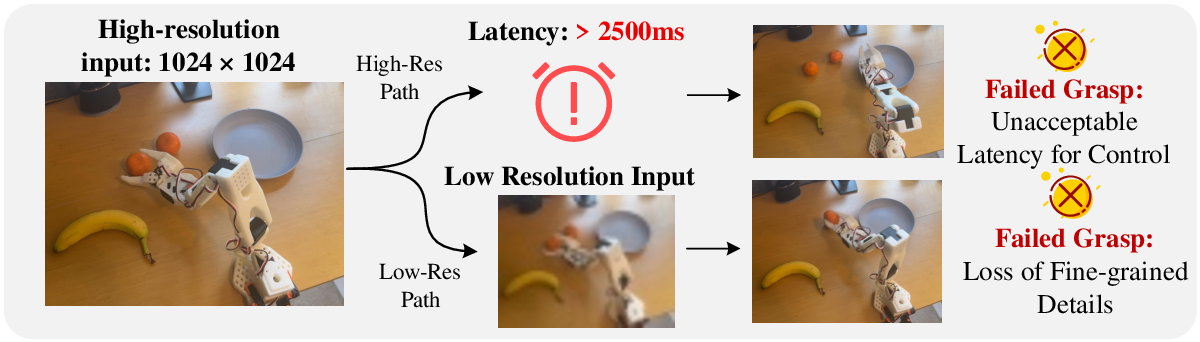}
        \caption{Existing VLA Dilemma (Latency vs. Resolution)}
        \label{fig:overview_a}
    \end{subfigure}
    
    \vspace{0.5em}
    
    % -------- (b) ----------
    \begin{subfigure}{\columnwidth}
        \centering
        \includegraphics[width=\columnwidth]{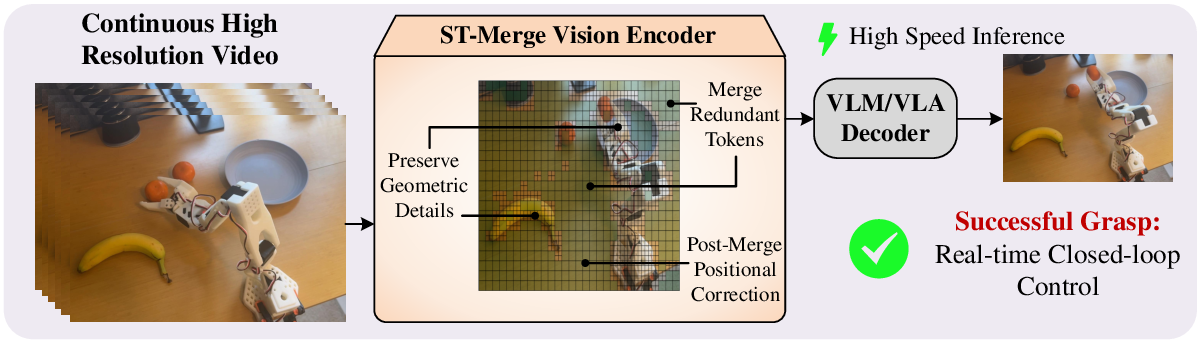}
        \caption{Our ST-Merge (Fast \& Precise)}
        \label{fig:overview_b}
    \end{subfigure}
    
    \caption{
    ST-Merge resolves the VLA dilemma between visual resolution and inference latency.
    By efficiently merging spatio-temporal tokens, it enables real-time, high-precision
    closed-loop robotic control.
    }
    \label{fig:overview}
    \vspace{-4mm}
\end{figure}

However, the deployment of VLMs and VLAs~\cite{kim2024openvla, intelligence2025pi_} on real physical robots is severely constrained by their extremely high inference latency.
As shown in Fig.~\ref{fig:overview}, Physical robots, on the other hand, require high-frequency, real-time closed-loop control to interact safely and effectively with dynamic environments.
The primary factor impeding such real-time deployment is the large volume of visual token inputs generated by continuous video streams.
The computational complexity of attention is $O(n^2)$, and dense visual tokens directly lead to exponential computing power consumption and unacceptable latency.
Therefore, existing VLMs and VLAs for real-time deployment often need to strike a balance between resolution and excessively long inference time.
For instance, state-of-the-art VLA models such as $\pi_{0.5}$ only support $244\times 244$ three-view images, comprising the top view and the wrist cameras on both arms.
This approach inevitably discards visual details in the images and cannot support continuous frame inputs, fundamentally limiting the performance of the models.

% Need Write More
Optimizing redundant vision tokens while preserving key semantic information has become a key challenge in VLM acceleration.
Existing methods~\cite{lee2024video, cao2023pumer, bolya2022token, lin2025boosting} have made some progress in vision token optimization.
However, most existing methods~\cite {lee2024video, chen2024image, lin2025boosting} optimize vision tokens within the LLM decoder or after the vision encoder, but the bottleneck in inference speed lies in the vision encoder. ToMe~\cite{bolya2022token} and PuMer~\cite{cao2023pumer} only compare token similarity, overlooking the spatial relationships among tokens.
FrameFusion~\cite{fu2025framefusion} relies solely on importance-based pruning, ignoring redundant features introduced by adjacent frames and repeated information.

% \begin{figure}[t]
% \centering
% \includegraphics[width=.5\textwidth]{Image/figure1.eps}
% \caption{ST-Merge resolves the VLA dilemma between visual resolution and inference latency. By efficiently merging spatio-temporal tokens, it enables real-time, high-precision closed-loop robotic control.}
% \label{fig:ST_Pipline}
% \end{figure}

To alleviate the real-time inference burden caused by the large-scale vision tokens generated by high-resolution and multi-frame visual inputs, we propose Spatio-Temporal Visual Token Merging (ST-Merge).
A plug-and-play, training-free spatio-temporal visual token merging framework that efficiently fuses redundant tokens during the visual encoding phase.
ST-Merge explicitly introduces spatio-temporal positional awareness into the token-matching process and constructs interpretable three-dimensional coordinates from the visual encoder's feature grid. 
By leveraging the feature similarity from the vision encoder, it performs a spatio-temporally consistent vision token merging.
We further introduced a post-merge positional correction that re-estimates rotary positional embeddings using the weighted centroid of source tokens, preserving geometric consistency and stabilizing long-range reasoning.

We evaluated the performance of ST-Merge by deploying it on a representative large-scale vision-language model (Qwen2.5-VL)~\cite{bai2025qwen2} and a state-of-the-art VLA policy ($\pi_{0.5}$)~\cite{intelligence2025pi_}.
For continuous video processing on Qwen2.5-VL, ST-Merge achieves a 2$\times$ inference speedup by reducing 70\% of the visual tokens, with a marginal accuracy drop of merely 1\%.
When applied to continuous robotic control on $\pi_{0.5}$, ST-Merge successfully shatters the latency bottleneck associated with high-resolution visual inputs. 
Evaluated on the LIBERO~\cite{liu2023libero} simulation benchmark, it delivers an exceptional 8.3$\times$ inference acceleration at a $1024\times1024$ resolution while strictly maintaining zero performance degradation.

The main contributions are as follows:

\begin{enumerate}
    \item We propose ST-Merge, a plug-and-play, training-free framework that efficiently fuses redundant tokens directly during the visual encoding phase. ST-Merge fundamentally alleviates the real-time inference burden caused by high-resolution and multi-frame continuous visual inputs in robotic systems.
    \item We introduce explicit 3D spatio-temporal coordinates and multi-queue parallel matching to achieve geometrically consistent token fusion in $O(n)$ time complexity. Furthermore, a novel Post-Merge Positional Correction mechanism that dynamically re-estimates rotary positional embeddings using the weighted centroid of vision tokens, crucially preserving the geometric consistency required for spatial reasoning in robotics.
    \item ST-Merge achieves a 2$\times$ inference speedup with a marginal 1\% precision loss on the Qwen2.5-VL in Video QA tasks, but also unlocks an extreme 8.3$\times$ speedup for the $\pi_{0.5}$ VLA policy at $1024\times1024$ resolution. 
    Remarkably, this acceleration comes with negligible performance degradation in both LIBERO~\cite{liu2023libero} simulations and real-world robotic deployments.
    % \item We comprehensively validate ST-Merge on both Qwen2.5-VL and the state-of-the-art VLA policy $\pi_{0.5}$, multi-resolution VLA deployment, up to 8.3× at 1024×1024 with zero degradation on LIBERO, and real-world closed-loop robot experiments.
    % \item We comprehensively deploy and evaluate ST-Merge on both a mainstream VLM (Qwen2.5-VL) and a state-of-the-art VLA policy ($\pi_{0.5}$). Evaluated on the LIBERO robotic simulation benchmark, ST-Merge successfully shatters the high-resolution latency bottleneck, delivering an exceptional 8.3$\times$ inference acceleration at a $1024\times1024$ resolution while strictly maintaining zero performance degradation.
\end{enumerate}

%% file: Content/02_Related_Work.tex
\section{Related Work}
\label{sec:related_work}

\subsection{Vision Language Model for Robotics}
\label{subsec:Related_VLM}

In recent years, large vision language models (VLMs)~\cite{achiam2023gpt, yang2025qwen3, bai2025qwen2, liu2023llava} have made impressive progress in robot manipulation, path planning~\cite{chen2024advanced}, visual semantic localization, and related areas.
Among them are frameworks such as CLIPort~\cite{shridhar2022cliport}, a vision-based manipulation method that constructs a language-conditioned imitation-learning agent.
VIMA~\cite{jiang2023vima} demonstrates excellent combinatorial generalization capabilities through interleaved multimodal prompts.
To bridge the gap between language reasoning and physical observations, PaLM-E~\cite{driess2023palm} incorporates sensor modalities as inputs to a VLM, and proposes a foundational model for an embodied brain capable.
Recent research has further overcome the inherent spatial and physical limitations of standard VLMs. 
RoboSpatial~\cite{song2025robospatial} enhances the model's 3D geometric perception capabilities, while PhysVLM~\cite{zhou2025physvlmenablingvisuallanguage} incorporates the robot's kinematic reachability constraints. 
Furthermore, to ensure the reliability of realistic physical execution, hybrid frameworks such as VLMPC~\cite{zhao2024vlmpc} tightly integrate VLM-based action candidate sampling with classic model predictive control.
VLMs that rely on video or image input serve as low-frequency high-level planners. 
The large amount of visual input often leads to significant inference latency, fundamentally hindering their real-time, high-frequency processing of high-resolution continuous observation streams.
% ToDo: Explaining the shortcomings of existing methods, and then introducing my method in all three sections.

\subsection{Vision Language Action Model}
\label{subsec:Related_VLA}

To overcome the latency bottlenecks of multi-stage pipelines, VLAs have emerged to map visual streams and text instructions directly into low-level robotic control policies end-to-end. Representative works like OpenVLA~\cite{kim2024openvla} provide a powerful open-source baseline fine-tuned on diverse cross-embodiment datasets, demonstrating exceptional zero-shot generalization. Furthermore, recent state-of-the-art models such as SmolVLA~\cite{shukor2025smolvla} and $\pi_{0.5}$~\cite{intelligence2025pi_} leverage advanced architectures (e.g., flow matching) to achieve remarkable performance in dexterous, long-horizon tasks. However, deploying these VLAs on physical robots for high-frequency, real-time closed-loop control presents a severe ``latency-resolution'' dilemma. To maintain actionable inference speeds, current leading policies like $\pi_{0.5}$ are often forced to aggressively downsample input video streams to exceptionally low resolutions (e.g., $244 \times 244$). While this brute-force reduction achieves near real-time execution, it inherently discards fine-grained visual details and compromises spatial precision. For complex dexterous manipulation that requires millimeter-level accuracy, this loss of spatial awareness is catastrophic. Consequently, there is an urgent need for an efficient token processing framework that enables VLAs to retain high-resolution spatial perception without suffering from prohibitive inference latency.

\begin{figure*}[ht]
    \centering
    \includegraphics[width=0.9\textwidth]{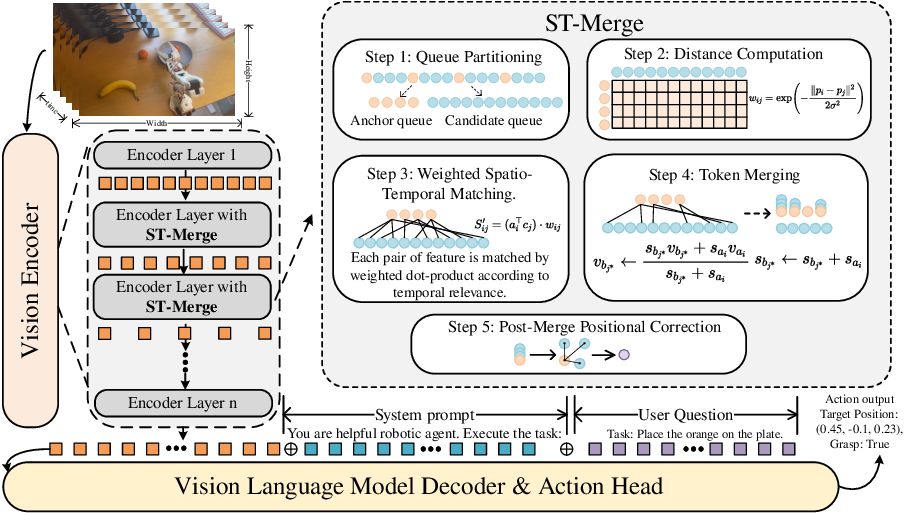}
    \caption{Pipeline of ST-Merge, which can be seamlessly deployed within existing LVLMs. Inserted into shallow layer of the vision encoder, ST-Merge (1) partitions tokens into an anchor queue and multiple candidate queues for parallel matching, (2) computes spatio-temporal neighborhood weights via Gaussian kernels, (3) performs weighted matching combining similarity and neighborhood priors, (4) merges tokens by confidence-guided, size-weighted averaging, and (5) applies RoPE-aware positional correction before feeding tokens to the language decoder.}
    \label{fig:pipline}
    \vspace{-4mm}
\end{figure*}

\subsection{Vision Token Reduction}
\label{subsec:Related_merge}

In VLMs, vision tokens are often dozens or even hundreds of times more than text tokens~\cite{sima2024drivelm, shao2025tr}, and they have the characteristics of spatial continuity and semantic sparseness. 
Vision Token Reduction primarily seeks to directly reduce the number of vision tokens via token merging~\cite{renggli2022learning, bolya2022token} and token pruning~\cite{kim2022learned, zhan2024exploring}. 
FastV~\cite{chen2024image} proposes a plug‑and‑play optimization scheme by learning adaptive attention patterns in the LLM’s early layers and pruning visual tokens in subsequent layers. 
VTW~\cite{lin2025boosting}, by contrast, retracts vision tokens in the deep layers of the LLM, retaining only text tokens while preserving performance. 
TopV~\cite{yang2025topv} formulates token pruning as an optimization problem, accurately identifying important visual tokens while remaining compatible with FlashAttention~\cite{dao2022flashattention}. 
However, all of these methods apply their optimizations only after the vision encoder has produced the vision embeddings, thereby overlooking the computational overhead incurred during the vision encoder stage of the VLM. 
ToMe~\cite{bolya2022token} introduces a Vision Transformer–based token‑merging method, but it has not been directly applied to VLMs. 
LLaVA-PruMerge~\cite{shang2024llava} reduces and merges visual tokens after the vision encoder by pruning low-attention tokens and clustering the remaining ones based on key similarity. 
Moreover, the latest VLMs~\cite{bai2025qwen2, liu2024deepseek} extend multimodal RoPE~\cite{heo2024rotary} to encode temporal and spatial positions better, whereas conventional visual token optimization methods, such as token merging, disrupt these positional encodings, preventing the VLM from correctly localizing merged tokens and thereby losing RoPE’s spatiotemporal awareness.

%% file: Content/03_Method.tex
\section{Method}

% \junzhou{Add a transitional content}

% \subsection{Overview}
Most existing methods~\cite{li2023llava, bai2025qwen2, chen2024internvl} optimize vision tokens only after the vision encoder or between language decoder layers. 
These methods can significantly improve speed when processing a single image containing a moderate number of tokens, but they struggle to scale to large-scale video vision token inputs. 
Furthermore, existing methods fail to incorporate the temporal dimension to optimize for temporally redundant vision tokens. 
Therefore, the key to accelerating large-scale language models with large vision token inputs lies in maintaining accuracy while reducing the computational overhead of the vision encoder.

To accelerate the video encoder, its computational load can be reduced by merging redundant vision tokens early in the network. 
ST-Merge is a training-free, plug-and-play token-merging strategy specifically designed for robot VLMs and VLAs. 
As shown in Fig.~\ref{fig:pipline}, ST-Merge is inserted into selected shallow layers of the vision encoder, before the multimodal adapter, performing spatio-temporal vision token merging before language decoding to reduce redundancy and accelerate inference. 
The overall pipeline consists of four key stages: Position-Aware Distance Computation (Sec.~\ref{subsec:position}); Weighted Spatio-Temporal Matching (Sec.~\ref{subsec:matching}); Token Merging and Feature Aggregation (Sec.~\ref{subsec:merge}); and Post-Merge Positional Correction (Sec.~\ref{subsec:positional}).

% The overall pipeline consists of four key stages: 
% Sec~\ref{subsec:position}: Position-Aware Distance Computation, 
% (2) Weighted Spatio-Temporal Matching, 
% (3) Token Merging and Feature Aggregation, 
% and (4) Post-Merge Positional Correction.

% \subsection{Spatio-Temporal Vision Token Merging}

\subsection{Position-Aware Distance Computation}
\label{subsec:position}

To achieve efficient vision token matching and merging in $O(n)$ time, and to avoid the ToMe~\cite{bolya2022token} bipartition strategy that causes vision tokens in the same queue to be unable to match, ST-Merge expands the dual-queue matching into a multi-partition queue, thereby achieving one-to-many parallel merge in a single token merge phase. Vision tokens are divided into $K$ sub-queues, where $G_1$ is the anchor queue and $G_{2:K}$ is the candidate queue. This allows each anchor token to simultaneously select the most similar neighbor from multiple candidate groups, thereby eliminating the serial bottleneck in traditional bipartition matching.

To better focus on spatio-temporal neighborhood information for vision token matching, ST-Merge constructs spatio-temporal feature-grid coordinates $p_i = (t_i, y_i, x_i)$ for each vision token within the vision encoder. Subsequently, the Euclidean distances between tokens are calculated, and relative position weights are generated based on a Gaussian kernel. The size of the neighborhood is controlled by $\sigma$:
\begin{equation}
    w_{ij} = \exp\!\left(-\frac{\|p_i - p_j\|^2}{2\sigma^2}\right),
\end{equation}

During the token matching phase, spatio-temporal neighborhood constraints are introduced to ensure that tokens are matched only within adjacent frames or spaces, rather than randomly matching globally. This explicit position encoding provides a position-aware prior for subsequent token matching, effectively avoiding mismatches across spaces or frames.

\subsection{Weighted Spatio-Temporal Matching}
\label{subsec:matching}
After obtaining the spatio-temporal neighborhood weights $w_{ij}$, the key vectors from the encoder layer are used to compute feature similarity. 
Given the anchor array $G_1=\{a_i\}$ and the candidate array $G_{2:K}=\{c_j\}$, the weighted similarity matrix is computed as follows:

\begin{equation}
S'_{ij} = (a_i^\top c_j) \cdot w_{ij},
\end{equation}
where $a_i^\top c_j$ represents semantic similarity, and $w_{ij}$ is the local neighborhood weight generated by the Gaussian kernel. This spatially weighted matching approach ensures consistency in the spatial and semantic features of high-scoring vision tokens, thereby preventing erroneous cross-region merging. For each anchor token $a_i$, its most similar candidate was selected:

\begin{equation}
    b_{j^*} = \arg\max_j S'_{ij}, \quad 
    s_i = \max_j S'_{ij}.
\end{equation}

All anchors are sorted in descending order by score $s_i$, and the top $r$ high-confidence matches are selected as merge candidate pairs. This matching process is similar to soft bipartite matching, with a complexity of only $O(n)$. It significantly improves the spatio-temporal consistency of token matching while maintaining matching efficiency.

% \textbf{Token Merging and Feature Aggregation.}
% Each source token $a_i$ is fused into the feature representation of the target token $b_{j^*}$ using a weighted average based on token size:
% \begin{equation}
% v_{b_{j^*}} \leftarrow
% \frac{s_{b_{j^*}} v_{b_{j^*}} + s_{a_i} v_{a_i}}
% {s_{b_{j^*}} + s_{a_i}},
% \qquad
% s_{b_{j^*}} \leftarrow s_{b_{j^*}} + s_{a_i},
% \end{equation}
% where $v$ represents the feature vector, $s$ represents the semantic weight or spatial coverage of the token. The multi-queue parallel structure greatly increases the number of merges in a single vision token merge, enabling the model to perform large merges in the shallow layers of the vision encoder alone.

\subsection{Token Merging and Feature Aggregation}
\label{subsec:merge}
Each source token $a_i$ is merged into its target token $b_{j^*}$ through a size-weighted feature aggregation:
\begin{equation}
v_{b_{j^*}} \leftarrow
\frac{s_{b_{j^*}} v_{b_{j^*}} + s_{a_i} v_{a_i}}
{s_{b_{j^*}} + s_{a_i}},
\qquad
s_{b_{j^*}} \leftarrow s_{b_{j^*}} + s_{a_i},
\end{equation}
where $v$ denotes the token feature vector and $s$ indicates its semantic weight or spatial coverage. 
This weighted fusion preserves the feature magnitude and spatial semantics of the merged region. 
By executing multiple queues in parallel, the proposed multi-queue structure enables a substantially higher number of merges within a single forward pass, allowing large-scale token reduction to be completed in the shallow layers of the vision encoder alone.

\subsection{Post-Merge Positional Correction} 
\label{subsec:positional}

The vision encoder of Qwen2.5‑VL uses RoPE~\cite{su2024roformer} to rotate the query and key vectors via a set of precomputed sine and cosine coefficients, thereby implicitly encoding spatial coordinates within the dot‑product attention. 
This mechanism substantially enhances the model’s spatial reasoning capabilities. 
However, when multiple tokens are merged into a single token, simply inheriting the positional embedding of one source token introduces significant spatial bias.

ST-Merge addresses this issue by dynamically computing the weighted centroid of the merged tokens. Specifically, a merged token is placed at the spatial centroid of all its constituent source tokens, with weights proportional to the number of original patches represented by each token. The position re‑estimation proceeds as follows: 
After merging, the correspondence between merged tokens and their original source tokens is recorded. 
For each merged token $i$, let $S_{ij}$ denote the merge size (or contribution weight) of source token $j$. 
The weights are first normalized as $\hat{S}_{ij} = \frac{S_{ij}}{\sum_{j} S_{ij}}$
The sine and cosine embeddings of the merged token are then recomputed as weighted sums of the original embeddings:

\begin{equation}
    \cos_{\text{new},i} = \sum_j^{\hat{S}_{ij}} \cdot \cos_{\text{ori},j},\quad \sin_{\text{new},i} = \sum_j^{\hat{S}_{ij}} \cdot \sin_{\text{ori},j}.
\end{equation}
% \begin{equation}

% \end{equation}
Finally, since the weighted combination may alter the vector norm, we renormalize the resulting $(\cos_{\mathrm{new}, i}, \sin_{\mathrm{new}, i})$ pair to maintain RoPE’s rotational invariance.

\subsection{ST-Merge deployment strategy}
% The St-Merge deployment strategy differs slightly from ISBI, so I need to update it.
% \junzhou{The ST-Merge deployment strategy is different from ISBI, so I need to update it.}

\input{Table/table_compare}

Mainstream VLM~\cite{shang2024llava, liu2024deepseek, bai2025qwen2, cao2023pumer} typically employ windowed attention in the majority of layers to balance computational efficiency with the need for long‑range visual dependencies, resorting to global attention only in a few selected layers to capture cross‑window information. Under a windowed mechanism, however, token similarities across different windows cannot be directly compared, making it difficult to guarantee consistency in the representations of merged tokens. In contrast, global attention provides a unified semantic space for all tokens. 
Motivated by this observation, we deploy ST-Merge within the shallow global‑attention layers of the vision encoder of VLM. 

In contrast, the deployment strategy in the VLA scenario differs. $\pi_{0.5}$ visual encoder is based on SigLIP and employs a global attention mechanism, without relying on RoPE for positional rotation encoding. We do not enable post-merge positional correction, the strategies introduced in Sec.~\ref{subsec:positional}, in the VLA. Meanwhile, $\pi_{0.5}$'s visual input is single-frame with multi-view input, without involving temporal modeling of consecutive video frames or cross-frame redundancy compression. Therefore, in VLA, we insert ST-Merge only in the shallow layer of the visual encoder to reduce the number of tokens as early as possible, and during matching, we use only spatial neighborhood constraints, without computing temporal positional weights. Apart from the above differences, the core processes of token matching and merging remain consistent.

Specifically, at the $l$-th layer, when the original token count $N_{l,\mathrm{orig}}$ is reduced to $N_{l,\mathrm{red}}$, the computational savings at each subsequent layer $k$ each can be approximated as:
\begin{equation}
    \text{FLOPs}_{\text{saved}} = \sum_{k = l+1}^{L} \left[ \alpha_k \left( N_{k,\text{orig}}^2 - N_{k,\text{red}}^2 \right) \right].
\end{equation}

This deployment strategy maximizes the reduction of computational overhead in subsequent layers while preserving long‑range interactions. 

%% file: Table/table_compare.tex
\begin{table*}[ht!]
\centering
\caption{Comparison of ST-Merge with the baseline and prior token-reduction methods under a unified evaluation setting. All approaches are evaluated on the same backbone. ST-Merge achieves the greatest reduction in FLOPs (51.04\% of the baseline) and the lowest inference latency, while preserving competitive average accuracy, indicating a favorable trade-off between efficiency and multimodal reasoning capability.}
\resizebox{\textwidth}{!}{
\begin{tabular}{lcccccc}
\hline
\textbf{Method}                   & \textbf{TFLOPs}                   & \textbf{Inference Speed (ms)} & \begin{tabular}[c]{@{}c@{}}\textbf{Robot Object} \\ \textbf{Interactions}\end{tabular} & \begin{tabular}[c]{@{}c@{}}\textbf{Temporal} \\ \textbf{Reasoning}\end{tabular} & \begin{tabular}[c]{@{}c@{}}\textbf{Intuitive} \\ \textbf{Physics}\end{tabular} & \textbf{Average}        \\ \hline
Qwen-2.5-VL 7B (Baseline)   & 55.07 (100\%)            & 1043.61              & 58.55                                                                & 63.12                                                         & 54.53                                                        & 58.73          \\
ToMe~\cite{bolya2022token}                     & 41.58 (75.50\%)          & 848.52               & 53.25                                                                & 58.00                                                         & 49.00                                                        & 53.41          \\
FastV~\cite{chen2024image}                    & 46.96 (85.27\%)          & 719.10               & 54.60                                                                & 60.25                                                         & 52.20                                                        & 55.68         \\
TempMe~\cite{shen2024tempme}                   & 36.79 (66.81\%)          & 605.77               & 54.55                                                                & 61.70                                                         & 49.80                                                        & 55.35          \\
\textbf{ST-Merge (ours)} & \textbf{28.11 (51.04\%)} & \textbf{414.90}      & \textbf{56.55}                                                       & \textbf{62.82}                                                & \textbf{53.69}                                               & \textbf{57.68} \\ \hline
\end{tabular}
}
\label{tab::main_result}
\end{table*}

%% file: Content/04_Experiment.tex
\section{Experiments}
\label{sec:experiments}

We concentrate on several experiments to answer the following research questions: 
\textbf{RQ1:} How does ST-Merge compare with other vision token reduction methods on continuous long-form video streams?
% How does ST-Merge perform compare to other 
% How does ST-Merge perform compared to other vision-token reduction methods on continuous long-video streams?
\textbf{RQ2:} How do the core components of ST-Merge contribute to overall performance?
\textbf{RQ3:} How does ST-Merge perform when deployed on VLA with different input resolutions?
\textbf{RQ4:} Can ST-Merge be effectively deployed in real-world robotic systems?

% \junzhou{Add more content to explain how we address those RQs.}

\label{research_question}

\subsection{Experiments Setup}
\label{subsec:exp_setup}

\subsubsection{Implementation details}
In this paper, we deploy ST-Merge on Qwen2.5-VL~\cite{bai2025qwen2} to verify whether the method enables VLM to process consecutive frame inputs quickly and accurately.
We deploy ST-Merge on $\pi_{0.5}$~\cite{intelligence2025pi_} to verify the VLA's performance under different resolution view inputs and in real-world deployments.
All model inference validation is performed on a single NVIDIA RTX 5090 host machine.
% For real-world VLA fine-tuning, we trained $\pi_{0.5}\; \text{base}$ model~\cite{lerobot_pi05_base} for 50,000 steps on NVIDIA H00 GPU.
We fine-tuned the $\pi_{0.5}\; \text{base}$ model~\cite{lerobot_pi05_base} for 50,000 steps on an NVIDIA H100 GPU across multiple setups, including LIBERO~\cite{liu2023libero} environments at $512 \times 512$ and $1024 \times 1024$ resolutions, as well as real-world scenarios.

\subsubsection{Evaluation Details}
We used the mini set from the MVP dataset~\cite{krojer2025shortcut} as the evaluation benchmark for Video VQA. The test scenarios primarily covered robot-object interactions, temporal reasoning, and intuitive physics and collisions.
In the LIBERO~\cite{liu2023libero} environment, we integrated ST-Merge into the $\pi_{0.5}$ model and deployed it on the LIBERO platform~\cite{liu2023libero} for testing. 
Finally, we designed three real-world grasping experiments to evaluate the grasping accuracy of the integrated ST-Merge at $512\times 512$ input resolution.

\begin{figure}[t!]
    \centering
    \includegraphics[width=\linewidth]{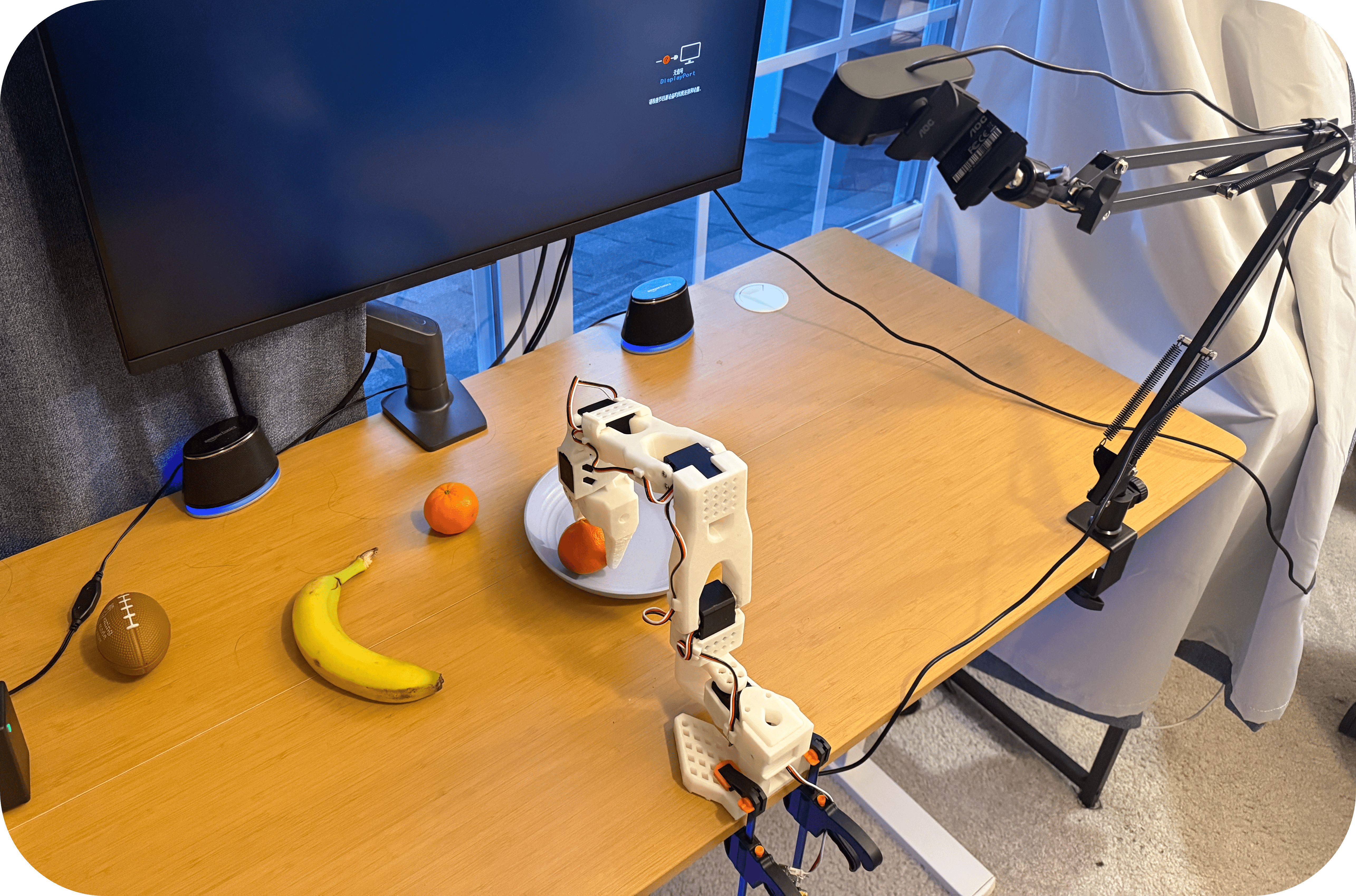}
    \caption{Real-World setup. The image shows the SO-101 arm and the camera used.}
    \label{fig:real_world}
    \vspace{-4mm}
\end{figure}

% \vspace{6mm}

\subsubsection{Real-World Dataset}
% Add more information about how we collect data
To further verify the effectiveness of deploying ST-Merge VLA on a real robot. 
As shown in Fig.~\ref{fig:real_world}, we collected three real-world robotic arm datasets based on the SO-ARM101~\cite{cadene2024lerobot}.
The datasets correspond to three representative manipulation tasks: 
(1) Move the banana to the left; 
(2) Place the orange on the plate; 
(3) Put the AirPods by the orange.
Specifically, a human operator controlled the SO-ARM101 leader arm while observing video captured from a fixed top-view camera.
The leader arm’s motions were mirrored by the SO-ARM101 follower arm in real time, 
ensuring precise alignment between visual observations and low-level control signals. 
For each, we collected 100 independent execution trajectories, resulting in a total of 300 real-world demonstrations.

\begin{figure*}[ht]
    \centering
    \includegraphics[width=0.9\textwidth]{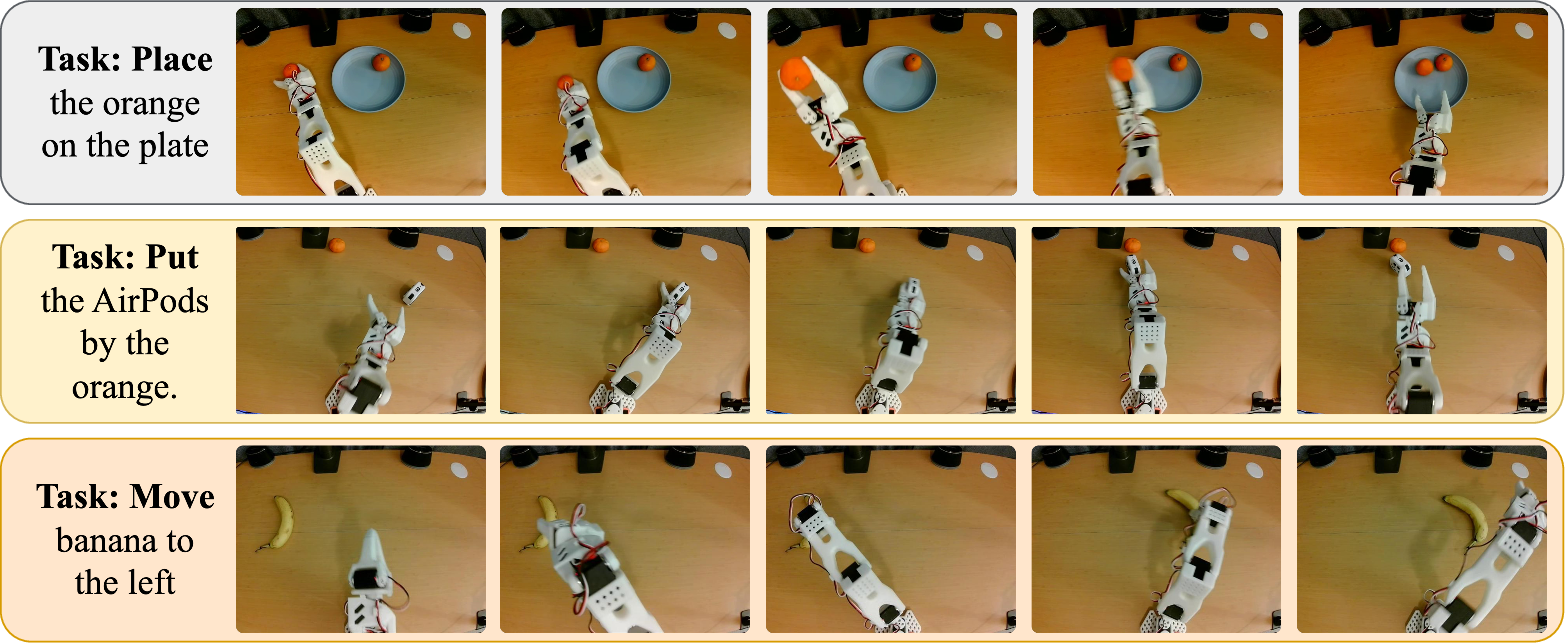}
    \caption{Qualitative results of real-world robotic manipulation on the SO-ARM101~\cite{cadene2024lerobot} platform, using the $\pi_{0.5}$~\cite{intelligence2025pi_} VLA policy equipped with ST-Merge at a $512 \times 512$ input resolution.
    The model successfully executes various tasks with high precision and low latency. 
    The spatially constrained token merging preserves fine-grained geometric cues, enabling accurate closed-loop control in real-world manipulation tasks.e}
    \label{fig:real_experiment}
\end{figure*}

\input{Table/table_libero}

% \subsection{Video QA on VLM}
\subsection{Evaluating Video QA on VLM}
\label{subsec:exp_VLM}
\subsubsection{Comparative study}

To answer \textbf{RQ1}, we compare ST-Merge against the original Qwen2.5-VL and representative token-reduction methods, including ToMe~\cite{bolya2022token}, FastV~\cite{chen2024image}, and TempMe~\cite{shen2024tempme}. 
We report accuracy, TFLOPs, and inference speed (ms).
As shown in Table~\ref{tab::main_result}, ST-Merge achieves the best end-to-end acceleration while preserving answer accuracy comparable to the baseline. 
Specifically, ST-Merge substantially reduces the overall computation and latency, yielding a 2$\times$ speedup, while requiring no training and no parameter modification.

The observed gaps across methods align with where token reduction is applied. 
FastV~\cite{chen2024image} performs decoder-side pruning based on cross-modal attention, which leaves the vision encoder cost largely unchanged and therefore often provides limited end-to-end speedup for high-token video inputs. 
ToMe~\cite{bolya2022token} reduces theoretical compute via per-layer merging, but repeated matching/merging can introduce non-trivial overhead. 
TempMe~\cite{shen2024tempme} incorporates spatio-temporal structure but relies on bipartite matching, which can be latency-heavy at scale. 
In contrast, ST-Merge reduces redundancy early in the vision encoder with low-overhead spatio-temporal merging and post-merge positional correction, leading to consistently lower latency.

\input{Table/table_ablation}

\subsubsection{Ablation study}

To answer \textbf{RQ2} and demonstrate that all steps are necessary, we conducted a step-by-step ablation study on Qwen2.5-VL-7B. 
As shown in Table~\ref{tab::ablation}, we started with only the basic ToMe module, which only performs token merging, and then added three modules: Queue Partitioning, Spatial Information Modeling, and Positional Correction. 

When applying only a basic merging strategy, computational cost and inference latency are reduced, but reasoning accuracy drops significantly. 
This indicates that traditional methods lack spatiotemporal awareness in continuous video settings, leading to the loss of critical local geometric information during merging.

To improve matching efficiency, we introduce Queue Partitioning. 
By transforming serial matching into multi-queue parallel processing, this mechanism substantially reduces inference latency and improves computational efficiency. 
However, accuracy does not recover accordingly, suggesting that improving matching speed alone is insufficient to ensure high-quality multimodal reasoning without structural constraints.

The addition of spatial information constraints leads to a clear performance improvement. 
By explicitly modeling spatiotemporal neighborhoods, this module imposes spatial boundaries on token matching and reduces erroneous cross-region merging, significantly enhancing accuracy with minimal computational overhead.

Incorporating Post-Merge Positional Correction enables the model to approach the original baseline performance. 
By dynamically adjusting RoPE, this module mitigates spatial distortions caused by changes in token granularity and restores the model’s ability to capture fine-grained geometric structures. 
In high-precision robotic manipulation tasks, such spatial consistency is essential.

\subsection{Closed-Loop VLA on LIBERO}
\label{subsec:exp_sim}

% \junzhou{ToDo: Retest the libero results. I find them a bit strange.}
% ToDo: Retest the libero results. I find them a bit strange.

% To answer \textbf{RQ3}, we evaluate ST-Merge in closed-loop control on the LIBERO benchmark by deploying it into a state-of-the-art VLA policy ($\pi_{0.5}$), as shown in Table~\ref{tab:libero}. 
For \textbf{RQ3}, we evaluate ST-Merge in closed-loop control on the LIBERO benchmark by integrating it into a state-of-the-art VLA policy ($\pi_{0.5}$), as shown in Table~\ref{tab:libero}.
High-resolution inputs (e.g., $1024\times1024$) are critical for fine-grained manipulation but incur quadratic attention cost, resulting in prohibitive latency, 2.6s per step for the baseline, which severely limits real-time control.

ST-Merge resolves this delay–resolution dilemma by intervening early in the visual encoder. 
Through spatio-temporal multi-queue matching, it aggressively merges static background redundancy prior to the computationally expensive decoding stage, ensuring that only motion-relevant and geometrically salient tokens are preserved. 
As resolution increases, ST-Merge avoids the typical computational explosion and achieves up to 8.3× speedup, reducing latency from seconds to 300ms at $1024\times1024$.

Crucially, this substantial compression does not degrade control performance: the success rate remains 97.5\%, matching the uncompressed baseline. These results demonstrate that ST-Merge enables real-time, high-resolution closed-loop control without sacrificing fine-grained physical reasoning.

\input{Table/table_real_world}

\subsection{Real-World VLA Deployment}
\label{subsec:exp_real}

To answer \textbf{RQ4}, we deploy our system on a real SO-ARM101 robotic platform and evaluate its closed-loop manipulation performance under real-world conditions.
Different from the simulation environment, a real-world robotic system faces a variety of uncontrollable factors. 
In this context, visual reasoning delay directly affects the temporal alignment between perception and action, thereby impacting the stability of closed-loop control.
We deploy the system on a physical robotic arm as shown in Fig.~\ref{fig:real_world}.
The execution process is shown in Fig.~\ref{fig:real_experiment}.
$\pi_{0.5}$ employs a cerebellum-cerebellum architecture: the cerebrum handles high-level visual understanding and action decision-making, while the cerebellum handles continuous control execution.
Therefore, whether the robotic arm lags or not does not accurately reflect the efficiency of reasoning. 
The key to the system's responsiveness lies in the refresh frequency of the cerebrum's decisions. 
We use the time required for each high-level decision as the core indicator, measured as ``thinking time.''

Without token optimization, the model's decision time for its first action is 582 ms, as shown in Table~\ref{tab:real_world}. 
This delay significantly reduces the effective control frequency and causes a time misalignment between perception and action. 
In actual execution, visual feedback lags during the grasping and alignment phase, resulting in pauses or oscillations in the robotic arm's trajectory. 
For example, in the Put task, failure to grasp AirPods often stems from millimeter-level pose errors. 
In the Place task, orange placement deviations are typically caused by a lag in depth information updates.

In contrast, integrating ST-Merge substantially improves closed-loop responsiveness by reducing visual processing overhead and restoring real-time control frequency. 
The system achieves consistently high success rates, as shown in Table~\ref{tab:real_world}, with smoother execution trajectories. 
This improvement stems from ST-Merge’s ability to aggressively remove background redundancy while preserving fine-grained geometric cues through spatio-temporal merging and positional correction.

%% file: Table/table_libero.tex
\begin{table*}[ht!]
\centering
% \scriptsize
\caption{Performance evaluation of the VLA policy $\pi_{0.5}$ with and without ST-Merge under varying visual resolutions. By integrating ST-Merge, the model achieves real-time closed-loop control speeds even at ultra-high resolutions without sacrificing the average manipulation accuracy.}
\resizebox{\textwidth}{!}{
\begin{tabular}{lcccccl}
\hline
\multirow{2}{*}{\textbf{Method}} & \multirow{2}{*}{\textbf{\begin{tabular}[c]{@{}c@{}}Inference Speed\\ (ms)(244$\times$244)\end{tabular}}} & \multirow{2}{*}{\textbf{Average Acc}} & \multirow{2}{*}{\textbf{\begin{tabular}[c]{@{}c@{}}Inference Speed\\ (ms)(512$\times$512)\end{tabular}}} & \multirow{2}{*}{\textbf{Average Acc}} & \multirow{2}{*}{\textbf{\begin{tabular}[c]{@{}c@{}}Inference Speed\\ (ms)(1024$\times$1024)\end{tabular}}} & \multirow{2}{*}{\textbf{Average Acc}} \\
                                 &                                                                                                   &                                       &                                                                                                   &                                       &                                                                                                     &                                       \\ \hline
Pi 0.5 (Baseline)                & 241.21                                                                                            & 95.83\%                                & 607.14                                                                                            & 96.67\%                                & 2614.56                                                                                             & 97.50\%                                \\
\textbf{ST-Merge (ours)}         & \textbf{141.55}                                                                                   & 94.16\%                                & \textbf{194.32}                                                                                   & 94.16\%                                & \textbf{312.11}                                                                                     & 97.50\%                                \\ \hline
\end{tabular}
}

\vspace{-4mm}

\label{tab:libero}
\end{table*}

%% file: Table/table_ablation.tex
\begin{table}[h]
\centering
% \scriptsize
\caption{Ablation study on Qwen2.5‑VL‑7B with cumulative addition of Token Merge, Queue Partitioning, Spatio Information, and Positional Correction. Numbers in parentheses denote percentage relative to the baseline.}
\resizebox{.5\textwidth}{!}{
\begin{tabular}{lccc}
\hline
\textbf{Method}         & \textbf{TFLOPs} & \textbf{Inference Speed (ms)} & \textbf{Acc}   \\ \hline
Qwen 2.5-VL (Baseline)  & 55.07           & 1043.61          & 58.73\%          \\
+ ToMe                  & 41.58           & 848.52           & 53.41\%        \\
+ Queue Partitioning    & 27.90           & 381.22           & 53.78\%        \\
+ Spatio Information    & 28.01           & 401.22           & 56.90\%        \\
+ Positional Correction & \textbf{28.11}  & \textbf{414.90}  & \textbf{57.68\%} \\ \hline
\end{tabular}
}
\label{tab::ablation}
% \vspace{-4mm}

\end{table}

%% file: Table/table_real_world.tex
\begin{table}[h]
\centering
% \scriptsize
\caption{Task success rates (\%) and average thinking time (ms) for $\pi_{0.5}$ with and without ST-Merge on move, place, and pick tasks. Each task is tested 20 times.}
\resizebox{.5\textwidth}{!}{
\begin{tabular}{lcccc}
\hline
\textbf{Method}              & \textbf{Place} & \textbf{Put} & \textbf{Move} & \textbf{Thinking Time (ms)} \\ \hline
 w/o ST-Merge & 80\%          & 70\%           & 50\%          & 582.14                        \\
w/ ST-Merge    & 80\%          & 75\%           & 55\%          & \textbf{191.22}                        \\ \hline
\end{tabular}

}
\label{tab:real_world}
\vspace{-4mm}
\end{table}

%% file: Content/05_Conclusion.tex
\section{Conclusion and Discussion}

% Conclusion Part
In this paper, we presented ST-Merge, a plug-and-play and training-free spatio-temporal visual token merging framework designed to resolve the latency–resolution dilemma in large VLMs and VLAs for real-time robotic control. 
By introducing explicit 3D spatio-temporal positional awareness, multi-queue parallel matching, and a RoPE-aware post-merge positional correction mechanism, ST-Merge efficiently reduces redundant visual tokens directly within the vision encoder while preserving geometric consistency. 
Extensive experiments on Qwen2.5-VL and the state-of-the-art VLA policy $\pi_{0.5}$ demonstrate significant acceleration—up to 8.3× speedup at 1024$\times$1024 resolution—without performance degradation on LIBERO and real-world manipulation tasks. 
These results show that high-resolution perception and real-time closed-loop control can be achieved simultaneously in embodied foundation models.

% Discussion Part
It is worth noting that our method can significantly improve speed while maintaining accuracy in high-resolution scenarios. 
This is because high-resolution inputs contain richer redundant information, and token merging primarily removes duplicate representations without compromising key semantics. 
Conversely, in low-resolution scenarios, the amount of original visual information is relatively limited, and further token merging may result in the loss of details, leading to a decrease in accuracy in simulation environments.

In real-world robot control, excessively long thinking time in VLA can cause visual feedback lag, thereby disrupting high-frequency updates in closed-loop control and manifesting as pauses or even oscillations in the robotic arm's trajectory. 
ST-Merge significantly reduces high-level decision latency, restoring the system's high-frequency closed-loop control capability; in some complex tasks, the overall success rate is even further improved due to the more timely and stable decision-execution link.